%% file: acl_latex.tex
\pdfoutput=1

\documentclass[11pt]{article}
\usepackage{hhline}
\usepackage[final]{acl}
\usepackage{booktabs}
\usepackage[export]{adjustbox}
\usepackage{array}
\usepackage{times}
\usepackage{latexsym}

\usepackage[T1]{fontenc}

\usepackage[utf8]{inputenc}

\usepackage{microtype}
\input{math_commands}

\title{MultiQG-TI: Towards Question Generation from Multi-modal Sources}

\author{Zichao Wang\thanks{$\,\,\,$Work done while at Rice University.} \\
  Adobe Research \\
\texttt{jackwa@adobe.com} \\\And
  Richard G. Baraniuk\\
  Rice University\\
  \texttt{richb@rice.edu} \\}

\begin{document}
\maketitle
\begin{abstract}
\vspace{-5pt}
We study the new problem of automatic question generation (QG) from multi-modal sources containing images and texts, significantly expanding the scope of most of the existing work that focuses exclusively on QG from only textual sources. We propose a simple solution for our new problem, called MultiQG-TI, which enables a text-only question generator to process visual input in addition to textual input. Specifically, we leverage an image-to-text model and an optical character recognition model to obtain the textual description of the image and extract any texts in the image, respectively, and then feed them together with the input texts to the question generator. We only fine-tune the question generator while keeping the other components fixed. On the challenging ScienceQA dataset, we demonstrate that MultiQG-TI significantly outperforms ChatGPT with few-shot prompting, despite having hundred-times less trainable parameters. Additional analyses empirically confirm the necessity of both visual and textual signals for QG and show the impact of various modeling choices. Code is available at {\small\url{https://rb.gy/020tw}}
\end{abstract}

\section{Introduction}
\vspace{-5pt}
Automatic question generation has the potential to enable personalized education experiences for subjects such as reading comprehension at a large scale~\cite{Wolfe1976,Kokku2018,Zhang2022,2206.04187} and improve standardized tests by reducing the costs and the test length~\cite{burstein2021theoretical}. Most, if not all, existing question generation (QG) methods operate {\it only on text}: they take a {\it textual} paragraph~\cite{Wang2018} or story~\cite{Xu2022} as input and generate a {\it textual} question. 
These methods' focus on text-based QG is limiting, because many interesting questions can involve, or be generated from, multiple modalities such as images, diagrams, and tables, in addition to texts~\cite{lu2022learn}. 

\begin{figure}[t!]
    \centering
    \includegraphics[width=\linewidth]{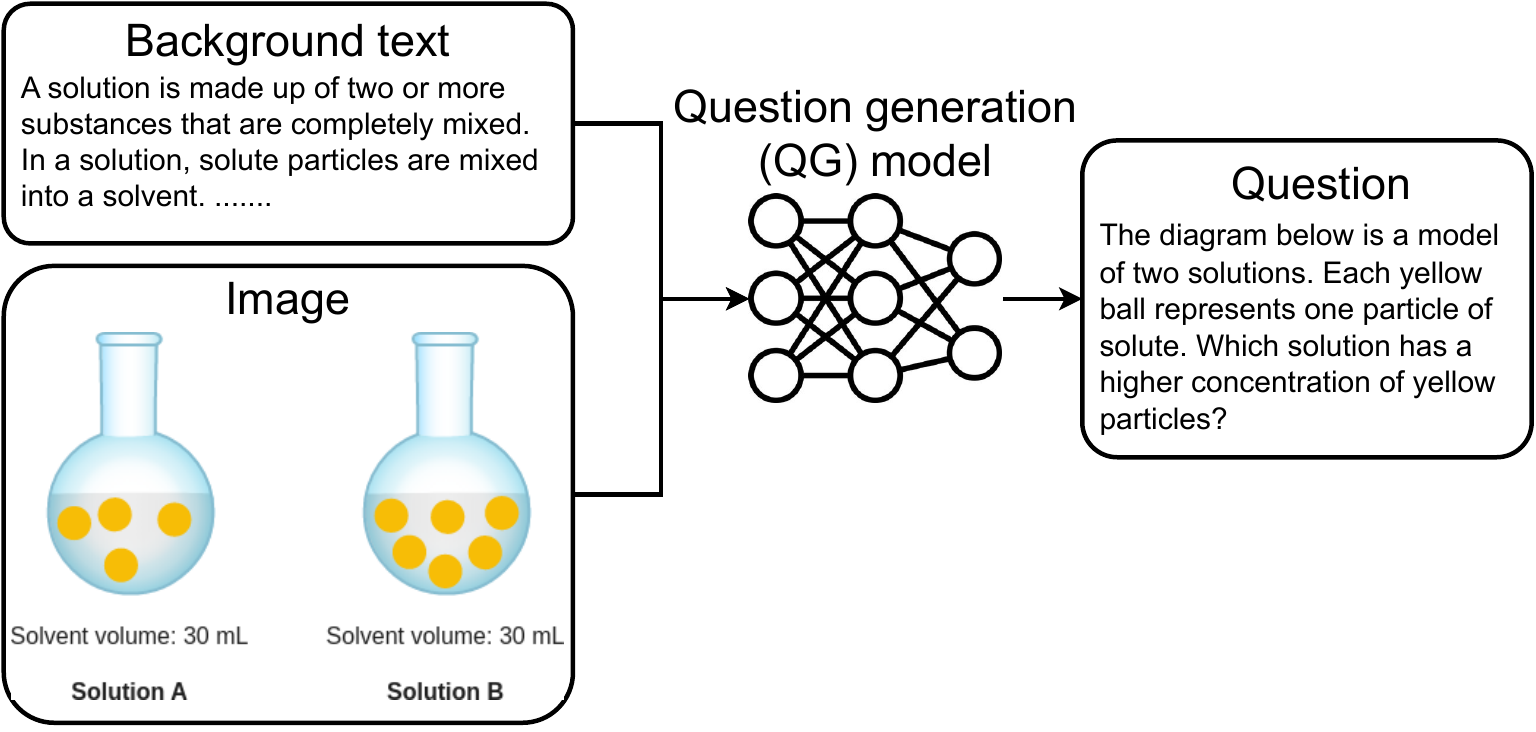}
    \vspace{-20pt}
    \caption{Illustration of our multi-modal question generation (QG) problem. Given a background text and an image, our goal is to develop a model to automatically generate a textual question based on them.}
    \vspace{-10pt}
    \label{fig:illustration}
\end{figure}

\begin{figure*}[t!]
    \centering
    \includegraphics[width=\linewidth]{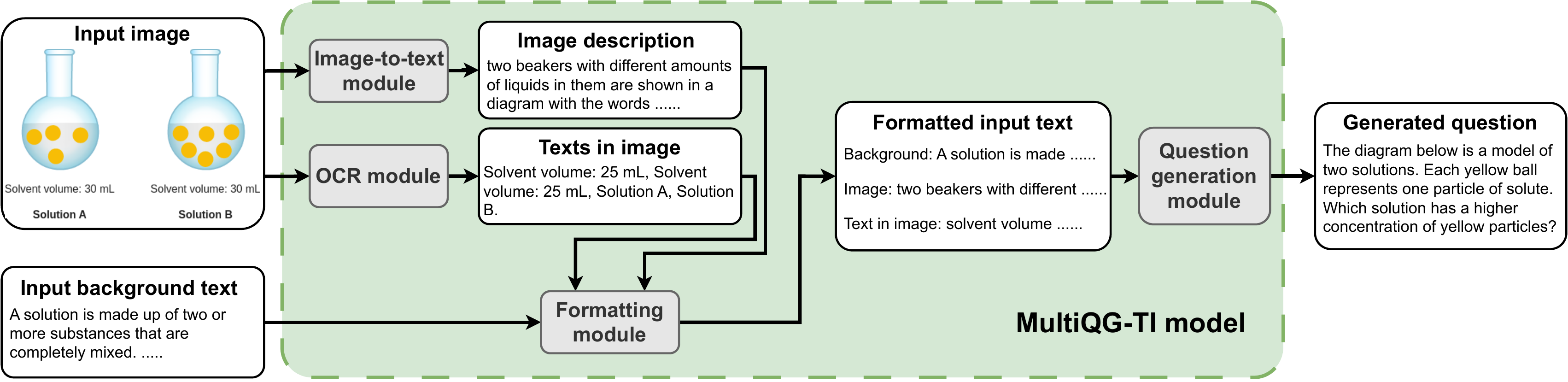}
    \caption{Illustration of the proposed MultiQG-TI methodology. 
    }
    \vspace{-10pt}
    \label{fig:model}
\end{figure*}

\vspace{-3pt}
\subsection{Contributions}
\vspace{-3pt}
In this paper, we conduct, to our knowledge, the {\it first} investigation into the under-explored problem of {\it multi-modal} question generation (QG). Specifically, we study the following problem: given multi-modal inputs containing both visual (e.g., an image) and textual (e.g., a textbook paragraph) information, we would like a model to output a textual question based on such multi-modal input. Note that the definition of visual input is very broad, e.g., it can be an image, a diagram, or a table in the image format. Although this multi-modal setting (image and text as input and textual question as output) is only a specific instance of multi-modality (one could consider using audio and video as input to generate questions, or generating questions with images in addition to texts), we argue that our setting is sufficiently broad and educationally meaningful. For example, many science questions ask about scientific phenomena, processes, and relationships commonly described in figures, diagrams, and tables~\cite{talmor2021multimodalqa,lu2022learn}.
We believe that our problem setting, illustrated in Figure~\ref{fig:illustration}, is an important first step toward more general multi-modal QG.

We propose a novel method, dubbed MultiQG-TI, for generating textual questions from multi-modal inputs of texts and images. The idea is simple: we enable a text-based question generator to ``see'' {\it by feeding it visual information in the form of text}. Specifically, we first use an off-the-shelf image-to-text model and an optical character recognition (OCR) model to produce a textual description of the image and extract the texts in the image. We then fine-tune a text-based generative model to generate a question given the input text and the text extracted from the input image. 
These components are readily available and require no or minimal fine-tuning, making MultiQG-TI easy to use and efficient to train.
Figure~\ref{fig:model} presents a high-level overview of MultiQG-TI.

We demonstrate MultiQG-TI's strong performance on the challenging ScienceQA dataset~\cite{lu2022learn}. For example, MultiQA-TI outperforms models using only texts or only images as input, demonstrating the necessity of including both texts and images as input in QG. MultiQA-TI also significantly outperforms ChatGPT in the few-shot in-context learning setting, demonstrating its competitiveness against much larger models. Finally, we analyze the factors that impact MultiQA-TI's performance, including the choices of image-to-text models and the sizes of the question generator model. We also provide generation examples to illustrate our method's strengths and errors.

\subsection{Related Work}
\paragraph{Question generation (QG) for education.}
QG models are often an integral component in personalized learning, intelligent tutoring systems, and assessment platforms to cheaply and scalably generate customized questions for each student~\cite{Le2014,1905.08949,Srivastava2021,white2022automated}.
For example, prior research has developed models to generate a variety of questions including those based on fairytales~\cite{Xu2022,Zhao2022}, factual questions~\cite{heilman-smith-2010-good,Wang2018}, and math word problems~\cite{Wang2021,Liu2021}.
Despite the rapid progress, most existing work focuses on {\it textual-based} QG. 
The exciting frontier of automatic multi-modal QG remains under-explored.

\paragraph{Multi-modal processing with text-only models.}
Our work is partially motivated by the recent line of work that demonstrate the possibility to use text-only models to perform visual-related tasks by feeding it text descriptors of the visual input. For example,~\citet{NEURIPS2022_381ceeae} enable large language models to perform video-related tasks such as event prediction by connecting them with image-to-text models. A few others take a similar approach to enable text-only models to perform captioning, reasoning, and question answering that involve videos or images~\cite{yang2022zeroshot,2303.11381,2211.09699}. 
However, the utility of their approach for multi-modal QG remains largely known.

\section{The MultiQG-TI Methodology}
We now describe the four modules in MultiQG-TI: a question generator module, an image-to-text module, an optical character reconigion (OCR) module, 
and an input formatting module.

\vspace{-2pt}
\paragraph{The question generator module.}
This module generates the question and
is the only trainable module in MultiQG-TI.
We adopt a text-based question generator such that its inputs must be all in text format. 
Adopting a text-based question generator enables us to choose from a wide range of pre-trained text-based generative models, whose training is also often more efficient than their multi-modal counterparts.
In this work, we instantiate the question generator with the recent Flan-T5 model~\cite{flan-t5} that have shown to perform strongly on new downstream tasks when fine-tuned on limited task-specific data.

\vspace{-2pt}
\paragraph{The image-to-text and OCR modules.}
A text-based question generator cannot take any visual input. To solve this problem, we use the image-to-text and OCR modules to interface between the image and text modalities and extract the visual information from the image format into a textual format appropriate as input for the text-based question generator. 
In particular, we use the image-to-text module to describe the {\rm content} in the image in texts, including any objects, scenes, actions, and events. We instantiate this module with the Flan-T5-XXL version of BLIP-2~\cite{2301.12597}.
While the image-to-text module extracts visually rich signals, it often fails to recognize any text in the image. This is problematic if the majority of the content in the image is text, such as a table. Therefore, we complement the image-to-text module with an OCR module that specializes in extracting the {\it texts} in the image. 
We instantiate the OCR module in MultiQG-TI with PaddleOCR~\cite{2009.09941}.

\vspace{-3pt}
\paragraph{The input formatting module.}
This module, $g$, is a simple function that concatenates the input text and the texts from the input image into one coherent textual input for the question generator model. There are many choices available and one can simply perform a string join operation. In this work, we apply input formatting with the following template: 
\texttt{\small Generate a question based on the following information. Background: \{input\_text\}. Image: \{image\_description\}. Texts in image: \{image\_text\}.}. In this template, \texttt{\small\{input\_text\}}, \texttt{\small\{image\_description\}}, and \texttt{\small\{image\_text\}} are placeholders that will be replaced with the actual input text, the output from the image-to-text model and the output from the OCR module, respectively.

\begin{table}[t]
\centering
\caption{MultiQG-TI (marked bold) significantly outperforms ChatGPT as well as variants with a single modality input across almost all metrics.}
\vspace{-7pt}
\label{table:main}
\begin{adjustbox}{max width=\linewidth}
\begin{tabular}{@{}lcccc@{}}
\toprule
\textbf{Method}  & \multicolumn{1}{l}{\textbf{BLEU}} & \multicolumn{1}{l}{\textbf{METEOR}} & \multicolumn{1}{l}{\textbf{ROUGE}} & \multicolumn{1}{l}{\textbf{BLEURT}} \\ \midrule
ChatGPT 0 shot           & 0.014                                & 0.264                               & 0.209                              & 0.448                               \\
ChatGPT 1 shot              & 0.021                                & 0.298                               & 0.208                              & 0.434                               \\
ChatGPT 3 shot              & 0.063                                & 0.332                               & 0.266                              & 0.449                               \\
ChatGPT 5 shot              & 0.088                                & 0.346                               & 0.301                              & 0.464                               \\
ChatGPT 7 shot              & 0.089                                & 0.342                               & 0.307                              & 0.460                                \\
\midrule
{\bf MultiQG-TI}                    & {\bf 0.725}                                & {\bf 0.829}                               & {\bf 0.830}                               & {\bf 0.757}                              \\ 
$\,\,\,$- text only                     & 0.570                                 & 0.714                               & 0.718                              & 0.675                               \\
$\,\,\,$- image only                &   0.714 &	0.817 &	0.813 &	0.760                \\\bottomrule
\end{tabular}
\end{adjustbox}
\vspace{-10pt}
\end{table}

\vspace{-3pt}
\paragraph{Training and inference.}
During training, we only update the parameters of the QG module while keeping the other modules fixed. We use the next word prediction as the training objective, which is commonly used in modern language model training~\cite{NIPS2017_3f5ee243}. During inference, we proceed as follows: given an input image and text, we first extract the text from the image using image-to-text module and the OCR module, then format them together with the input text, and finally feed the formatted texts to the fine-tuned QG module to generate a question.

\begin{table*}[t!]
\caption{An example of a question in physics generated by MultiQG-TI.}
\vspace{-7pt}
    \label{tab:example}
    \centering
    \begin{adjustbox}{max width=\linewidth}
    \begin{tabular}{p{1\textwidth}c}
        \toprule
        \centering{\textbf{Input background text}} & \textbf{Input image} \\
        \cmidrule(l{13em}r{13em}){1-1}\cmidrule(lr){2-2}
        Magnets can pull or push on other magnets without touching them. When magnets attract, they pull together. When magnets repel, they push apart. These pulls and pushes are called magnetic forces. Magnetic forces are strongest at the magnets' poles, or ends. Every magnet has two poles: a north pole (N) and a south pole (S). Here are some examples of magnets. Their poles are shown in different colors and labeled. Whether a magnet attracts or repels other magnets depends on the positions of its poles. If opposite poles are closest to each other, the magnets attract. The magnets in the pair below attract. If the same, or like, poles are closest to each other, the magnets repel. The magnets in both pairs below repel. & \raisebox{-.85\height}{\includegraphics[scale=0.18,valign=c]{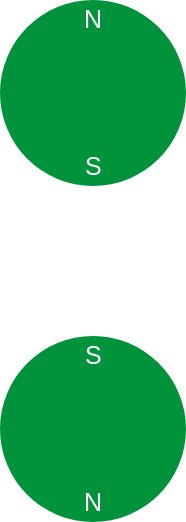}} \\
        \midrule
        \multicolumn{2}{c}{\textbf{MultiQG-TI generated question}} \\
         \cmidrule(l{16.5em}r{10em}){1-1}
        \multicolumn{2}{c}{Two magnets are placed as shown. Will these magnets attract or repel each other?}  \\
        \bottomrule
    \end{tabular}
    \end{adjustbox}
    \vspace{-10pt}
\end{table*}

\vspace{-3pt}
\section{Experiments}
\vspace{-3pt}
\paragraph{Dataset.} 
We use the ScienceQA dataset~\cite{lu2022learn} throughout our experiments, which we preprocess and split into training, validation, and test splits.
All results in this paper are reported on the test split.
More details on the dataset and preprocessing steps are in Appendix~\ref{app:data}.

\vspace{-3pt}
\paragraph{Baselines.}
Because there are no prior work on automatic multi-modal QG, we use off-the-shelf model APIs and variants of MultiQG-TI as the baselines. Specifically, we use {\bf ChatGPT} API~\cite{2203.02155} with zero-shot and in-context learning~\cite{2001.08361,wei2022emergent} with up to seven examples, each of which is formatted exactly the same as our preprocessed data points in the ScienceQA dataset. We also compare with {\bf MultiQG-TI with only a single modality as input} (i.e., either only text or only image). 

\vspace{-3pt}
\paragraph{evaluation.}
We choose four evaluation metrics including {\bf BLEU}, {\bf METEOR}, {\bf ROUGE}, and {\bf BLEURT}, all of which have been widely used in existing QG works. We report all results, except for those using ChatGPT API, based on the average of 4 random, independent runs. More details on the experiment setup, baselines, and evaluation are in Appendices~\ref{app:model} and~\ref{app:baseline}.

\begin{figure}
    \centering
    \includegraphics[width=0.9\linewidth]{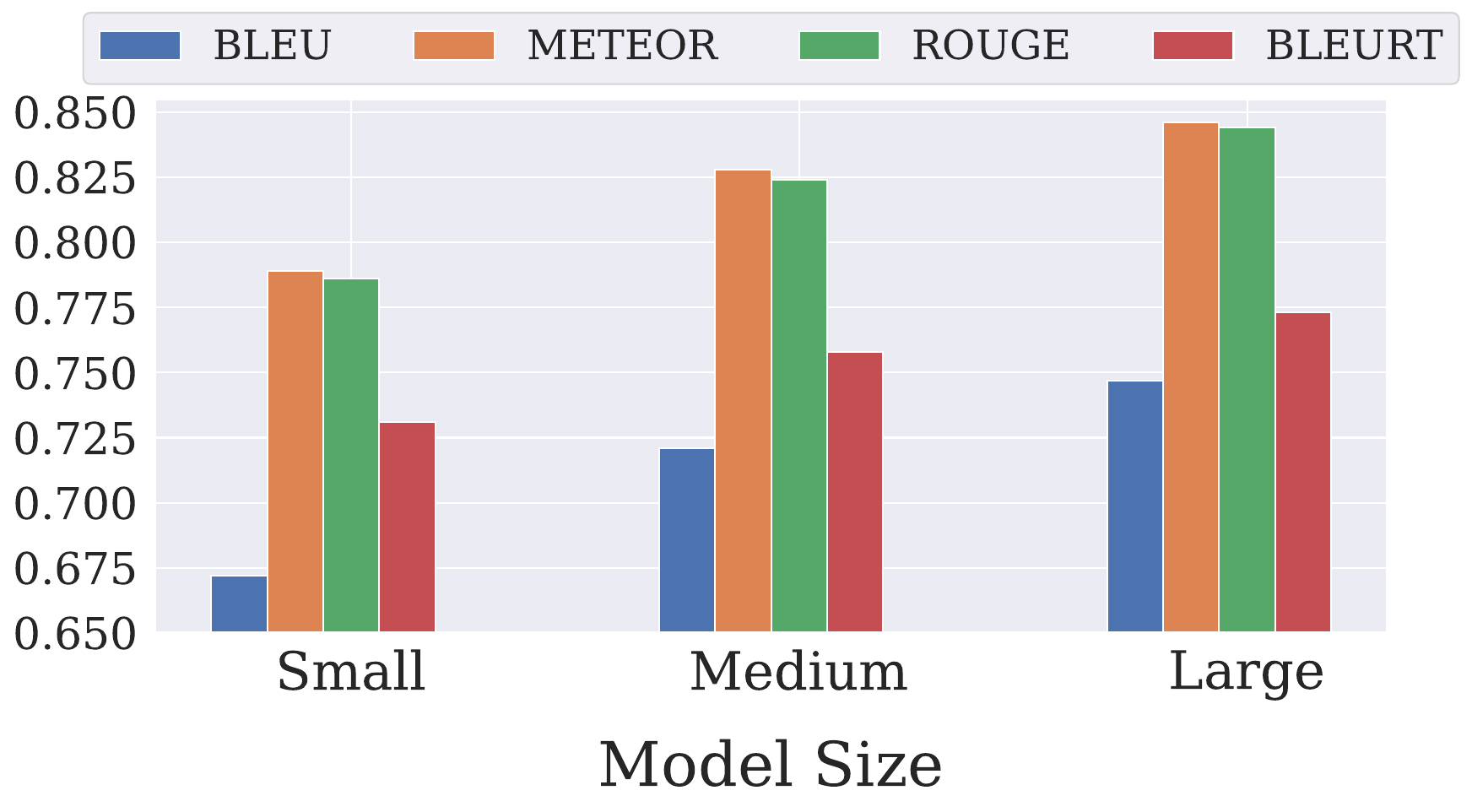}
    \vspace{-10pt}
    \caption{Larger model tends to result in improved QG performance across all metrics.}
    \vspace{-10pt}
    \label{fig:ablation_modelsize}
\end{figure}

\vspace{-2pt}
\subsection{Main quantitative results}
\vspace{-1pt}
Table~\ref{table:main} summarizes the main results.\footnote{For conciseness, we choose not to report standard deviations because all of them are quite small (around $0.002$).}
These results clearly show that ChatGPT fails at the multi-modal QG task in our setting. Although its performance steadily improves with more examples in the in-context learning setting, ChatGPT trails MultiQG-TI by a gigantic margin. The comparison between ChatGPT and MultiQG-TI reminds one to be cautious when using ChatGPT in specialized tasks such as multi-modal QG and presents strong empirical evidence that a small, fine-tuned model is still highly relevant in certain generation tasks. 
Table~\ref{table:main} also demonstrate the benefits of including both the visual and textual information when generating questions because MultiQG-TI outperforms its variants with only textual or only visual input.

\subsection{Analyses}
\vspace{-5pt}
\paragraph{The choice of question generators.}
We study the impact of the model size of the QG module on the QG performance and summarize the results in Figure~\ref{fig:ablation_modelsize}, where ``small'', ``medium'', and ``large'' represent the Flan-T5 variants of 80 million, 250 million, and 780 million parameters, respectively. The figure implies that a larger model generally leads to improved performance across all evaluation metrics. Notably, by fine-tuning only on a few thousand training examples with a modest-sized model, MultiQG-TI achieves high performance,\footnote{As a comparison, some of the latest QG works achieve a BLEURT score of up to 0.67; see the results of a recent QG competition: {\small\url{https://www.thequestchallenge.org/leaderboard}}} making it appealing for practical use and deployment in resource-constrained settings.

\begin{table}[t]
\centering
\caption{A sufficiently large image-to-text model leads to better QG performance, although the benefit of model size diminishes as the size increases beyond 2.7 billion parameters.}
\vspace{-7pt}
\label{table:ablation-vit-model}
\begin{adjustbox}{max width=\linewidth}
\begin{tabular}{@{}lcccc@{}}
\toprule
\textbf{ViT model}      & \textbf{bleu\_4} & \textbf{meteor} & \textbf{rouge} & \textbf{bleurt} \\ \midrule
ViT-GPT2 (239M)      & 0.671            & 0.79            & 0.785          & 0.733           \\
BLIP2-OPT (2.7b)   & 0.744            & 0.843           & 0.843          & 0.770            \\
BLIP2-OPT (6.7b)   & 0.743            & 0.842           & 0.841          & 0.773           \\ 
BLIP2-Flan-T5-XXL (11b) & {\bf 0.747}            & {\bf 0.846}           & {\bf 0.844}          & {\bf 0.773}           \\
\bottomrule
\end{tabular}
\end{adjustbox}
\vspace{-10pt}
\end{table}

\vspace{-5pt}
\paragraph{The choice of image-to-text models.}
We also study the impact of the image-to-text models on the QG performance and summarize the results in Table~\ref{table:ablation-vit-model}. 
Specifically, we compare BLIP2-Flan-T5-XXL (11 billion parameters), the image-to-text model we use in MultiQG-TI, to three smaller variants ranging from 239 million to 2.7 billion, and 6.7 billion parameters, respectively. 
We observe that QG performance improves steadily but minimally after the model becomes larger than 2.7 billion parameters, although the largest model still wins modestly. 
These results imply that MultiQG-TI may retain the same level of competitiveness even with a smaller off-the-shelf image-to-text model, suggesting more resource-saving opportunities without compromising performance.

\paragraph{Qualitative examples.}
We show an example generated question by MultiQG-TI in Table~\ref{tab:example}, as well as additional ones in Appendix~\ref{app:examples}. These examples further illustrates MultiQG-TI's capability in generating fluent, coherent, and meaningful questions from multi-modal scientific contexts. 
We also provide an in-depth analyses of the errors that MultiQG-TI makes during generation, which we defer to Appendix~\ref{app:examples} due to space constraint.

\section{Conclusion}
We have conducted a first study into automatic multi-modal QG from images and texts. Our proposed solution, MultiQG-TI, is simple, easy-to-use, and highly capable, as evaluated and analyzed on the ScienceQA dataset. Our work opens a myriad of research opportunities. Some of the exciting future directions include: 1) QG with multi-modal inputs and multi-modal outputs; 2) end-to-end vision-language modeling approach for QG; and 3) evaluating and comparing the pedagogical utilities of questions generated from multi-modal sources in real-world educational scenarios.

\section*{Acknowledgements}
This work was supported by NSF grants 1842378, ONR grant N0014-20-1-2534, AFOSR grant FA9550-22-1-0060, and a Vannevar Bush Faculty Fellowship, ONR grant N00014-18-1-2047.

\bibliography{anthology,custom}
\bibliographystyle{acl_natbib}

\newpage
\clearpage
\appendix

\section{Experiment details}
\label{app:exp-details}
\subsection{Dataset and preprocessing}
\label{app:data}
Each data point in the ScienceQA dataset contains the question text, a background text, and an image. The total number of data points in the ScienceQA dataset is 21,208. We refer readers to~\citet{lu2022learn} for more details on the dataset.  However, the background text and the image are optionally included. As a result, not all data points contain both the background text and the image. We only keep data points that contain all three elements, resulting in 5,942 data points. We further randomly split them into train, validation, and test splits, resulting in 3606/1204/1132 data points in the train/validation/test splits, respectively. For  both the remaining texts and images, we did not perform further processing and keep them as-is before feeding them to the MultiQG-TI components that are responsible for processing them.

We note that the MultimodalQA dataset~\cite{talmor2021multimodalqa} is also an appropriate dataset choice with rich multi-modal information beyond just texts and images.
Because our present work focuses on image and text as input modalities, we leave more complex data modalities for QG for future work.

\subsection{MultiQG-TI model details}
\label{app:model}
\paragraph{Image-to-text generation.}
We use contrastive sampling~\cite{su2022a} with the following parameters:\footnote{See this blog post for an explanation of the different parameters that appear in contrastive sampling: {\small\url{https://huggingface.co/blog/introducing-csearch}}} $\alpha=0.6$ and $k=4$, with a temperature of 1, n-gram penalty of 3, and minimum text description length of 30 tokens. For each given image, we sample 10 different text descriptions, rerank them by the image-to-text model's perplexity, and choose the best description (with the lowest perplexity score) as the final text description for the image, which we will then send to the QG module, together with the OCR module's output and the input background text.

\paragraph{QG module training.}
We perform all training on a single NVIDIA Quadro RTX 8000 GPU. For all QG module variants that we consider, we use the same training setup. Specifically, we train it with a learning rate of $0.0003$ for 8 epochs with early stopping if validation loss does not improve over the most recent 3 epochs. We use a batch size of 3 with a gradient accumulation step of 4, resulting in an effective batch size of 12 (e.g., the parameters are updated every 12 training steps). We also clip the gradients to 1 to stabilize training. All these training procedures are standard in training text generative models.

\paragraph{Inference and evaluation.}
We use the same contrastive sampling strategy as in image-to-text generation. Additionally, we sample 10 generated questions, rerank them by perplexity, and fetch the best-ranked sample as the final generated question for each input text-image pair in the test set. All evaluations are conducted on this ``top-1'' setting. For each individual run, we perform the above sampling strategy with a different seed to obtain a different set of generated questions for each input in the test set. We then perform the same evaluation on each generated set and then average the results, resulting in the averaged quantitative evaluations reported in the main paper.

\paragraph{Remarks.}
MultiQG-TI leverages readily available, open-source tools to solve the new problem of multi-modal question generation. 
Its modular design makes it flexible and easily adaptable, enabling one to upgrade a component when a more capable one becomes available.
Moreover, the only trainable component is the question generator. There are many choices available for this component, any of which can achieve competitive performance with relatively limited model sizes, making it suitable for low-resource training settings.
An end-to-end multi-modal QG model is still methodologically interesting and we leave this as a future work.

\subsection{ChatGPT baseline}
\label{app:baseline}
We use the \texttt{gpt-3.5-turbo-0301} model API throughout our experiments. The system message we give to the model at the beginning of the API call is as follows: \texttt{You are a helpful assistant. Your job is to generate a question, which consists of a question background/context and the question itself, given the user's provided context information, which consists of an instruction, background, subject, topic, and category. Your answer should be in the following template: 'Question context: ... Question: ...'}. After that, for zero-shot QG, we send the templated input background text, OCR extracted text from the input image, and the text description of the input image to the API, formatted exactly as what we would do for MultiQG-TI. For few-shot QG, we construct each example as a pair of input and output, where the input is the templated input consisting of the input text and texts extracted from the input image, and the output is the corresponding question text to the input text and image. We only perform generation once for each setting and for each input to avoid incurring higher costs of making OpenAI API calls.

\paragraph{Selecting examples for in-context learning.}
We perform a basic cosine similarity search for each input context and image pairs. Specifically, we first encode each formatted input text (recall, it contains the input background text, the image description, and the texts in the image) as a vector using the SentenceTransformers.\footnote{\small{\url{https://www.sbert.net/}}}. Then, for each formatted input in the test set, we perform a similarity search, computing its cosine similarity with every formatted input in the training set, and select up to seven most similar formatted input as the examples to be used in prompting ChatGPT in the few-shot in-context learning setting.

\section{Additional literature review}
The MultimodalQA dataset~\cite{talmor2021multimodalqa} actually involves a cursory description of generating questions from multiple sources. However, the QG process described therein relies on human annotation, a manual process that cannot achieve automatic QG and therefore is neither a baseline to our work nor related to our goal of automatic QG. 

Recent research has demonstrated the impressive capabilities of models that can connect data from multiple modalities, such as generating images from texts~\cite{2204.06125} and vice versa~\cite{He2017}.
Specifically related to our work, recent advances in vision-language models~\cite{alayrac2022flamingo,2301.12597,2303.08774} enable models to converse with a user given both texts and images. However, most demonstrated use cases of these models are in casual dialogues~\cite{2301.12597}, image captioning~\cite{Hossain2019}, and visual question answering~\cite{Antol_2015_ICCV}. The utilities of these models for QG remain largely unknown.

\section{Additional results}
\paragraph{Additional examples of generated questions.}
\label{app:examples}
We provide additional generation examples in Table~\ref{tab:app:example} for chemistry, physics, and biology, respectively. These examples corroborate with the one in the main text and demonstrate the capability of MultiQG-TI in generating reasonable questions from image and text inputs.

\begin{table*}[t!]
\caption{Additional examples of science questions generated by MultiQG-TI.}
\vspace{-5pt}
    \label{tab:app:example}
    \centering
    \begin{adjustbox}{max width=\linewidth}
    \begin{tabular}{p{0.7\textwidth}c}
        \toprule
        \centering{\textbf{Input background text}} & \textbf{Input image} \\
        \cmidrule(lr){1-1}\cmidrule(lr){2-2}
        A solution is made up of two or more substances that are completely mixed. In a solution, solute particles are mixed into a solvent. The solute cannot be separated from the solvent by a filter. For example, if you stir a spoonful of salt into a cup of water, the salt will mix into the water to make a saltwater solution. In this case, the salt is the solute. The water is the solvent. The concentration of a solute in a solution is a measure of the ratio of solute to solvent. Concentration can be described in terms of particles of solute per volume of solvent. concentration = particles of solute / volume of solvent & \raisebox{-1\height}{\includegraphics[scale=0.4,valign=c]{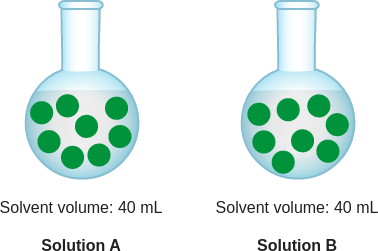}} \\
        \midrule
        \multicolumn{2}{c}{\textbf{MultiQG-TI generated question}} \\
        \cmidrule(l{16.5em}r{-3em}){1-1}
        \multicolumn{2}{c}{The diagram below is a model of two solutions. Each pink ball represents one particle of solute.}  \\
        \multicolumn{2}{c}{Which solution has a higher concentration of pink particles?} \\[10pt]
        \bottomrule \\[-10pt]
        \bottomrule \\
        \centering{\textbf{Input background text}} & \textbf{Input image} \\
        \cmidrule(lr){1-1}\cmidrule(lr){2-2}
        A force is a push or a pull that acts on an object. Each force acts on an object in a certain direction. If two forces act on an object in opposite directions, they are called opposing forces. & 
        \raisebox{-0.5\height}{\includegraphics[scale=0.25,valign=c]{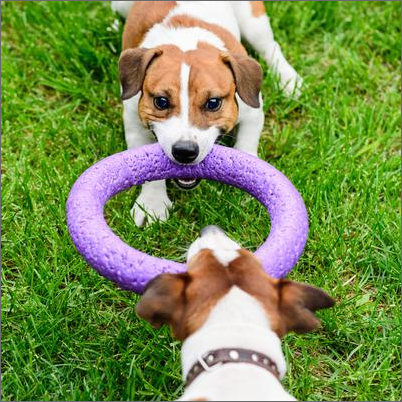}} \\
        \midrule 
        \multicolumn{2}{c}{\textbf{MultiQG-TI generated question}} \\
        \cmidrule(l{16.5em}r{-3em}){1-1}
        \multicolumn{2}{c}{The text below describes a pair of opposing forces. Opposing forces act on an object in opposite directions. Read the text.} \\
        \multicolumn{2}{c}{Then answer the question below. Two puppies, Rusty and Coco, play with a toy. Think about two of the forces that act on} \\
        \multicolumn{2}{c}{ the toy:  Coco pulls toward herself. Rusty pulls away from Coco. As Coco pulls on the toy, what is the direction of the} \\
        \multicolumn{2}{c}{ opposing force?} \\[10pt]
        \bottomrule \\[-10pt]
        \bottomrule \\
        \centering{\textbf{Input background text}} & \textbf{Input image} \\
        \cmidrule(lr){1-1}\cmidrule(lr){2-2}
        Offspring genotypes: homozygous or heterozygous?
How do you determine whether an organism is homozygous or heterozygous for a gene? Look at the alleles in the organism's genotype for that gene.
An organism with two identical alleles for a gene is homozygous for that gene.
If both alleles are dominant, the organism is homozygous dominant for the gene.
If both alleles are recessive, the organism is homozygous recessive for the gene.
An organism with two different alleles for a gene is heterozygous for that gene.
In a Punnett square, each box represents a different outcome, or result. Each of the four outcomes is equally likely to happen. Each box represents one way the parents' alleles can combine to form an offspring's genotype. 
Because there are four boxes in the Punnett square, there are four possible outcomes.
An event is a set of one or more outcomes. The probability of an event is a measure of how likely the event is to happen. ...... \texttt{(truncated due to excessive length)} & \raisebox{-1.2\height}{\includegraphics[scale=0.5,valign=c]{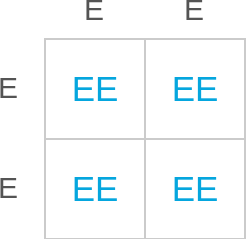}} \\
\midrule 
        \multicolumn{2}{c}{\textbf{MultiQG-TI generated question}} \\
        \cmidrule(l{16.5em}r{-3em}){1-1}
        \multicolumn{2}{c}{In a group of rainbow trout, some individuals have a greenish-brown body and others have a blue body. In this group, } \\
        \multicolumn{2}{c}{the gene for the body color trait has two alleles. The allele for a blue body (e) is recessive to the allele for a greenish-} \\
        \multicolumn{2}{c}{brown body (E). This Punnett square shows a cross between two rainbow trout. What is the probability that a } \\
        \multicolumn{2}{c}{ rainbow trout produced by this cross will be heterozygous for the body color gene?} \\
        \bottomrule

    \end{tabular}
    \end{adjustbox}
    \vspace{-10pt}
\end{table*}

\paragraph{Qualitative generation error analysis.}
MultiQG-TI is not without problems. In Table~\ref{tab:app:error}, we provide an exemplary erroneous generated question to illustrate the typical problems that MultiQG-TI has when performing QG. 

In our observation, there are two major sources of error. 
The first one comes from the mistakes cascaded from the image-to-text model. In the example in Table~\ref{tab:app:error}, the object in the image is dolerite, but the image-to-text model in MultiQG-TI recognizes it as granite, resulting in the image description ``a black piece of granite on a white background''. As a result, the question generator, which generates the question conditioned on the image description, picks up the wrongly reconigized object ``granite'' and use it to generate a question on granite instead of on dolerite. 

The second source of error comes from hallucination, a major bottleneck preventing language models from real-world, high-stake use scenarios~\cite{Ji2023}. MultiQG-TI is not immune to this problem. In the example in Table~\ref{tab:app:error}, the question generator produces the phrase ``pure substance'', which is neither a property of dolerite nor granite because both are mixtures. 

These are challenging issues to tackle. For example, it is even difficult for a non-expert to identify the object in the image in Table~\ref{tab:app:error}. Similarly, it is difficult to verify the factual correctness of the generated question without resorting to external sources such as web search and textbooks. Reducing these errors would require improvements to the image-to-text model and mitigating hallucination in language models, both of which remain active areas of research.

\begin{table*}[t!]
\caption{An illustrative erroneous question generated by MultiQG-TI. The texts marked in {\color{red}red} are the erroneous parts. The word ``{\color{red}granite}'' results from the image-to-text model, which incorrectly recognizes the dolerite in the image as granite. The phrase ``{\color{red}pure substance}'' is neither a property of dolerite nor granite.}
\vspace{-5pt}
    \label{tab:app:error}
    \centering
    \begin{adjustbox}{max width=\linewidth}
    \begin{tabular}{p{0.7\textwidth}c}
        \toprule
        \centering{\textbf{Input background text}} & \textbf{Input image} \\
        \cmidrule(lr){1-1}\cmidrule(lr){2-2}
        Minerals are the building blocks of rocks. A rock can be made of one or more minerals.
Minerals and rocks have the following properties:
Property | Mineral | Rock
It is a solid. | Yes | Yes
It is formed in nature. | Yes | Yes
It is not made by organisms. | Yes | Yes
It is a pure substance. | Yes | No
It has a fixed crystal structure. | Yes | No
You can use these properties to tell whether a substance is a mineral, a rock, or neither.
Look closely at the last three properties:
Minerals and rocks are not made by organisms.
Organisms make their own body parts. For example, snails and clams make their shells. Because they are made by organisms, body parts cannot be  minerals or rocks.
Humans are organisms too. So, substances that humans make by hand or in factories are not minerals or rocks.
A mineral is a pure substance, but a rock is not.
A pure substance is made of only one type of matter.  Minerals are pure substances, but rocks are not. Instead, all rocks are mixtures.
A mineral has a fixed crystal structure, but a rock does not.
The crystal structure of a substance tells you how the atoms or molecules in the substance are arranged. Different types of minerals have different crystal structures, but all minerals have a fixed crystal structure. This means that the atoms and molecules in different pieces of the same type of mineral are always arranged the same way.
However, rocks do not have a fixed crystal structure. So, the arrangement of atoms or molecules in different pieces of the same type of rock may be different!
 & \raisebox{-1.5\height}{\includegraphics[scale=0.7,valign=c]{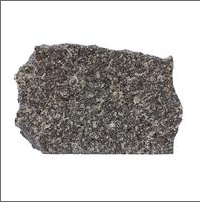}} \\
        \midrule
        \multicolumn{2}{c}{\textbf{MultiQG-TI generated question}} \\
        \cmidrule(l{16.5em}r{-3em}){1-1}
        \multicolumn{2}{c}{{\color{red}Granite} has the following properties: {\color{red}pure substance} solid fixed crystal structure naturally occurring not made by}  \\
        \multicolumn{2}{c}{living things Question: Is granite a mineral or a rock?} \\
        \bottomrule
    \end{tabular}
    \end{adjustbox}
    \vspace{-10pt}
\end{table*}

\end{document}

%% file: math_commands.tex
\usepackage{amsmath,amsfonts,bm}

\def\eqref#1{equation~\ref{#1}}

\def\1{\bm{1}}

\def\vc{{\bm{c}}}

\def\vy{{\bm{y}}}

\def\mI{{\bm{I}}}

\DeclareMathAlphabet{\mathsfit}{\encodingdefault}{\sfdefault}{m}{sl}
\SetMathAlphabet{\mathsfit}{bold}{\encodingdefault}{\sfdefault}{bx}{n}

